\documentclass{article}
\usepackage{ijcai26}

\usepackage{times}
\usepackage{soul}
\usepackage{url}
\usepackage[hidelinks]{hyperref}
\usepackage[utf8]{inputenc}
\usepackage[small]{caption}
\usepackage{graphicx}
\usepackage{amsmath}
\usepackage{amsthm}
\usepackage{booktabs}

\usepackage{algorithm}
\usepackage{algorithmic}
\usepackage[switch]{lineno}
\usepackage{xcolor}
\usepackage{amssymb}
\usepackage{tabularx}
\usepackage{multirow}
\usepackage{placeins}
\usepackage[inline]{enumitem}
\usepackage{dblfloatfix}
\usepackage{afterpage}
\usepackage{makecell}
\usepackage[table]{xcolor}
\title{Cross‑Relational Preference Learning for Better LLM Instruction Following

}

\author{
Runsheng Li$^{1,2}$
\and
Kai Sun$^{1,2}$\thanks{Corresponding author.}\and
Bin Shi$^{1,2}$\And
Bo Dong$^{2,3}$\\
\affiliations
$^1$School of Computer Science and Technology, Xi’an Jiaotong University, Xi'an, Shaanxi 710049, China\\
$^2$Shaanxi Provincial Key Laboratory of Big Data Knowledge Engineering, 
Xi'an Jiaotong University, Xi'an, Shaanxi 710049, China\\
$^3$School of Distance Education, Xi'an Jiaotong University, Xi'an, Shaanxi 710049, China\\
\emails
lirunsheng@stu.xjtu.edu.cn,
\{sunkai, shibin, dong.bo\}@xjtu.edu.cn
}
\begin{document}

\maketitle

% , thereby significantly boosting instruction‑following performance

\begin{abstract}
    Large Language Models (LLMs) still exhibit limited capability in following complex instructions. While existing approaches often rely on preference learning to enhance this ability, they typically overlook the relationships between the permissible response spaces of different instructions, which restricts a model to align with subtle and diverse constraint variations. To address this, we propose Cross-Relational Preference Learning (CRPL), a novel framework for constructing preference data that explicitly models inter-instruction relationships through two key techniques: \textit{Cross-Relationship Perturbation} and \textit{Cross-Region Pair Sampling}. This enables the generation of more diverse preference data that captures a wide spectrum of constraint variations. Additionally, we introduce an atomic constraint-based verification mechanism to rigorously assess response satisfaction, ensuring high-quality preference pair construction. Extensive experiments across multiple preference learning methods (e.g., DPO, KTO), LLM backbones and four instruction-following benchmarks demonstrate that our approach achieves substantial improvements over prior baselines and exhibits strong generalization.
\end{abstract}

\section{Introduction}

Instruction following is a core capability of large language models (LLMs), requiring precise understanding of user instructions and the generation of high-quality responses that adhere to instruction constraints~\cite{survey,metric}. Although advanced LLMs \cite{llama3herdmodels,deepseekr1,yang2025qwen3technicalreport} excel at handling simple instructions, their performance remains limited when faced with complex instructions involving multiple constraints \cite{ifbench,pham-etal-2024-suri}.

To enhance LLM performance on complex instructions, existing methods employ reinforcement learning to align outputs with constraints~\cite{fcs,Mastery}. A central challenge lies in constructing high-quality positive and negative samples for preference optimization. Early approaches focused on instruction-level optimization, synthesizing complex instructions and using verifiers to perform rejection sampling—where passing and failing responses form preference pairs~\cite{ultraif,autoif}. 

\begin{figure}[t]
  \includegraphics[width=\columnwidth]{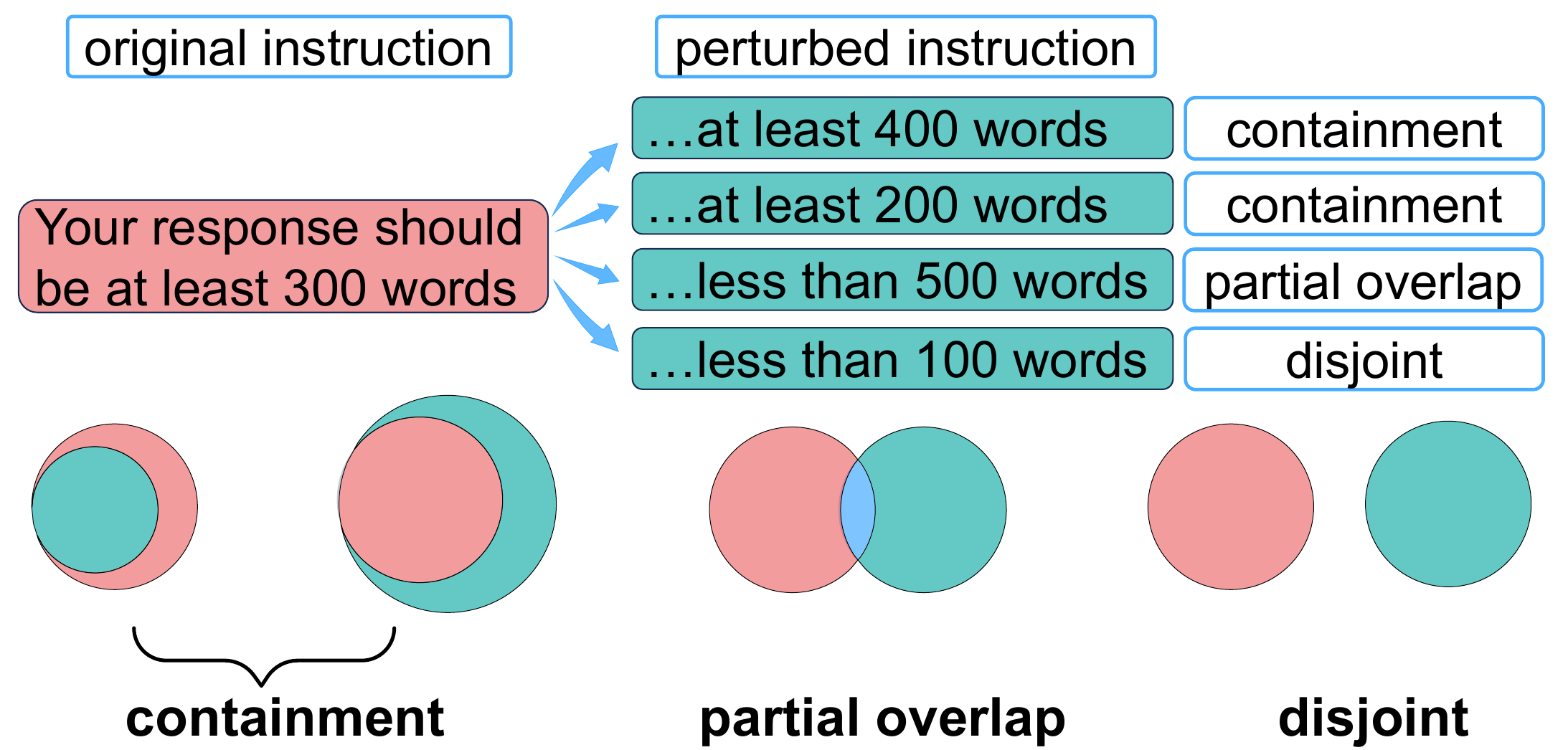}
  \caption{Relationships between the permissible response spaces of an original instruction and its perturbed variant. Colors denote: \textcolor{red}{red} — original instruction's response space; \textcolor[rgb]{0.137,0.694,0.675}{green} — perturbed instruction's response space; \textcolor[rgb]{0.282,0.678,1}{blue} — intersection of the two spaces.}
  \label{fig1}
\end{figure}

Recent works have shifted toward constraint-level optimization, aiming to enhance the model’s understanding of individual constraints within instructions. Specifically, several methods generate positive responses by correcting only the violated portions of negative responses, thereby refining the model's comprehension of specific constraints~\cite{spar,bmc}. Alternatively, some methods introduce random perturbations, such as deletion, modification, or addition, to specific constraints in the instructions. The model then learns from the differences between responses generated before and after perturbation~\cite{iopo,huang-etal-2025-musc}. 

Despite promising results, existing approaches suffer from a critical limitation: they fail to adequately account for the relationships between the permissible response spaces of different instructions when constructing preference pairs. We classify these relationships into three fundamental types: containment, partial overlap, and disjoint, as illustrated in Figure~\ref{fig1}. Existing methods either neglect these relationships~\cite{conifer,autoif} or focus solely on the disjoint case~\cite{huang-etal-2025-musc,iopo}. Consequently, the resulting preference data lack diversity, restricting the model's ability to learn from comprehensive variants of constraints. To address this limitation, we propose that training should incorporate more diverse preference pairs that explicitly capture these inter-space relationships, analogous to how humans leverage relational concepts such as similarity and opposition for comparative learning.

% 's ability

% enhance a model's adherence to intricate instructions. Our framework follows a perturbation paradigm, first decomposing complex instructions into atomic, verifiable constraints. Then perturbations are applied to each atomic constraint to generate variant instructions. The core innovation of our approach lies in

Based on these insights, we propose Cross-Relational Preference Learning (CRPL), a novel framework to enhance LLMs' adherence to intricate instructions. CRPL follows a perturbation paradigm, but it explicitly models the relationships between the response spaces of the original and perturbed instructions. This is achieved through two key designs:

% Our perturbation strategy is explicitly guided by the three fundamental types of relationships. Specifically, 

(1) \textbf{Cross-Relationship Perturbation:} When modifying a specific constraint, we steer the perturbation direction according to a target relationship. For instance, to construct a containment relationship, we instruct the model that ``\textit{the perturbed valid-response set must be a proper superset of the original valid-response set}''. This approach enables the model to generate comprehensive constraint variants that span all three relationship types.

% (1) \textbf{Relationship Guided Perturbation:} When modifying a constraint, we explicitly guide the model to control the direction of perturbation, informed by a specific relationship. For instance, to construct a containment relationship, we instruct that ``\textit{the perturbed valid-response set must be a proper superset of the original valid-response set}''. This allows the model to systematically induce desired relationships between response spaces.

(2) \textbf{Cross-Region Pair Sampling:} For a given relationship, we ensure that the sampled positive responses collectively cover all pertinent regions within the combined response space of the original and perturbed instructions. As illustrated for the partial-overlap relationship in Figure~\ref{fig1}, the positive responses for the original and perturbed instructions span the disjoint regions (red and green) as well as their intersection (blue). This ensures that the constructed positive-negative pairs capture the full spectrum of the relationship, thereby providing diverse and representative data for subsequent preference optimization.

% Region-Across Pair Sampling requires that 

% significantly improving verification accuracy under intricate instructions

% thereby  within an instruction

To ensure that positive responses are strictly drawn from specific regions of the combined response space, we introduce an atomic constraint-based verification mechanism. This mechanism constructs and applies a separate verification function for each atomic constraint, providing a rigorous guarantee on the quality of the resulting preference pairs. Unlike existing methods, they typically rely on an LLM to judge the entire instruction-response pair or to generate a single verification function (e.g., as in UltraIF~\cite{ultraif} and IOPO~\cite{iopo}), which often fail to adequately verify all constraints.

% To ensure that the positive responses be strictly drawn from specific regions of the combined response space, we introduce an atomic constraint-based verification mechanism. By constructing and applying a separate verification function for each atomic constraint, this method significantly improves verification accuracy under intricate instructions. In contrast, existing methods rely on an LLM to judge the entire instruction-response pair or generate a single verification function~\cite{ultraif,iopo}, which often fail to verify all constraints adequately.

Our contributions can be summarized as follows:
\begin{itemize}
    \item We propose Cross‑Relational Preference Learning (CRPL), a novel framework that enhances the diversity of preference pairs by explicitly modeling the relationships between response spaces.
    \item  We introduce an atomic constraint‑based verification mechanism to rigorously assess whether sampled responses satisfy a given instruction, thereby ensuring the construction of high‑quality preference data.
    \item Through extensive experiments across multiple preference learning methods (e.g., DPO, KTO), diverse LLM backbones and four instruction‑following benchmarks, we demonstrate that our framework consistently improves instruction‑following performance and exhibits strong generalization.
\end{itemize}
\section{Related Work}

\subsection{Preference Learning}
Preference learning is an important paradigm for enhancing the instruction-following capabilities of LLMs. Early work typically collected human preference data to train reward models and then optimized LLMs using reinforcement learning such as PPO \cite{ppo} to ensure adherence to human preferences \cite{instructgpt}.
Subsequently, preference learning methods such as DPO \cite{dpo}, KTO \cite{KTO}, and SimPO \cite{SimPO} have been proposed to directly optimize the model’s output distribution to better align with human preferences, without explicitly training a reward model.

Recently, some works have adapted the DPO algorithm to enhance models' ability to discern fine-grained distinctions between instructions. For example, \cite{bmc} and \cite{huang-etal-2025-musc} propose token-level preference optimization methods to achieve fine-grained instruction alignment. IOPO \cite{iopo} proposes an alignment method that simultaneously considers preferences over both inputs and outputs. In addition, some methods employ an iterative DPO training strategy to achieve continuous improvement in instruction-following capabilities \cite{spar,ultraif}. More recently, several works \cite{ifbench,lambert2025tulu,verif} have leveraged Reinforcement Learning with Verifiable Rewards (RLVR) to enhance models’ ability to follow complex instructions by utilizing verifiable reward signas.

\subsection{Preference Data Construction for Instruction-Following}
Constructing high-quality preference data is critical for preference learning~\cite{survey}. A simple yet effective approach is rejection sampling~\cite{reject1,llama3herdmodels}. For instance, ULTRAIF \cite{ultraif} trains an instruction composer to iteratively synthesize complex constrained instructions and evaluate responses to obtain preference data. AutoIF \cite{autoif} leverages LLMs to generate instructions and code to verify responses. 

Alternatively, some methods construct preference data by correcting negative responses to obtain corresponding positive responses. For example, From Complex to Simple (FCS) \cite{fcs} uses a student model to generate initial responses, which are then corrected by a teacher model to form preference pairs. SPAR \cite{spar} employs tree-search-based stepwise self-correction on instruction-violating responses to produce highly comparable preference pairs. BMC \cite{bmc} generates highly correlated preference pairs by applying minimal modifications to negative responses. Other methods generate contrasting responses by perturbing the original instructions. IOPO \cite{iopo} perturbs certain constraints in the original instruction to induce responses that oppose the original constraints, forming preference pairs. MUSC \cite{huang-etal-2025-musc} decomposes each complex instruction into atomic constraints and randomly removes some constraints to create negative instructions, with responses from the original and negative instructions forming preference pairs.

% in a key aspect. Specifically, we improve existing methods

Our work contributes to the construction of high-quality preference data but differs from prior approaches by explicitly modeling the relationships between the response spaces of original and perturbed instructions. This approach enhances the diversity of preference pairs, enabling the model to learn from comprehensive constraint variations and thereby significantly improving its instruction-following capability.

\section{Problem Definition}
Complex Instruction Following is a conditional text generation problem subject to a set of specific constraints.

\paragraph{Notation.}
Let $\mathcal{X}$ denote the space of complex instructions. Given an input instruction $x_j \in \mathcal{X}$, we posit that $x_j$ entails a set of distinct constraints $\mathcal{C}_j = \{c_1, c_2, \dots, c_m\}$, where $m$ is the number of constraints in $x$. Each constraint $c_i$ represents a specific requirement (e.g., length limit, keyword inclusion, or formatting style) that must be satisfied.

\begin{figure*}[t]
  \includegraphics[width=\textwidth]{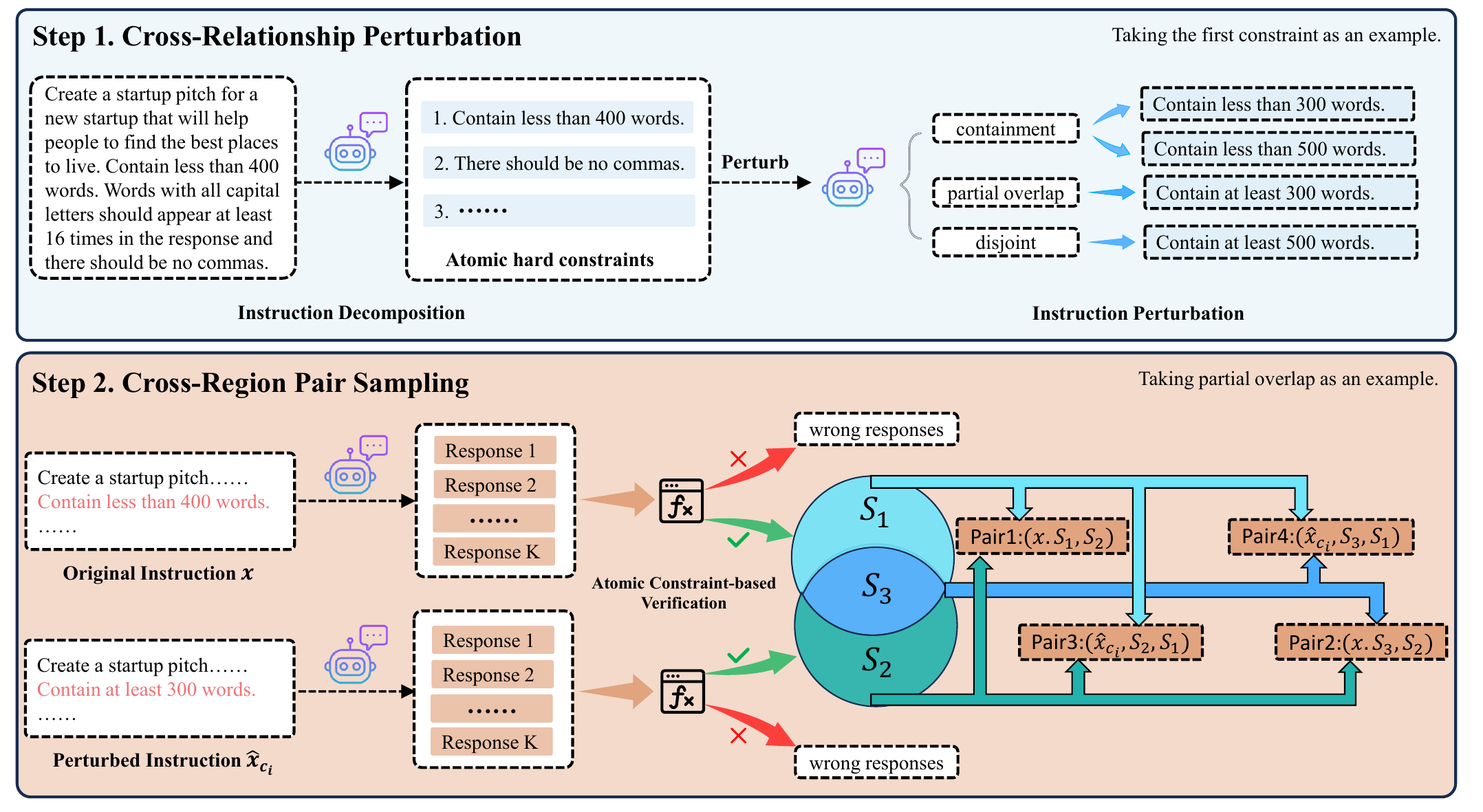}
  \caption{An overview of our Cross-Relational Preference Learning (CRPL) framework. (1) Cross-Relationship Perturbation: We decompose the original instruction into atomic constraints and generate perturbed instructions that exhibit specific response-space relationships (containment, partial overlap, or disjoint) with the original one. (2) Cross-Region Pair Sampling: Guarded by an atomic constraint-based verification mechanism, we sample positive and negative responses from distinct sub-regions of the combined response space to construct diverse, high-quality preference pairs.}
  \label{fig2}
\end{figure*}

\paragraph{Task Formulation.} 
The goal is to generate a response $y_j$ that follows all constraints in $\mathcal{C}_j$. Formally, we aim to learn a model parameterized by $\theta$ that approximates:
\begin{equation}
\begin{aligned}
    \hat{y} = \mathop{\arg\max}_{y} P_\theta(y \mid x) \quad \\ \text{s.t.} \quad \mathbb{V}(y, c_i) = 1, \forall c_i \in \mathcal{C}_x
\end{aligned}
\end{equation}
where $\mathbb{V}(\cdot, \cdot)$ is a verification function such that $\mathbb{V}(y, c_i) = 1$ if the response $y$ satisfies constraint $c_i$, and $0$ otherwise.

\section{Method}
Our framework consists of two key steps: Cross-Relationship Perturbation and Cross-Region Pair Sampling. First, we guide a strong LLM to generate perturbations for a given instruction across three predefined relationship types (i.e., containment, partial overlap or disjoint). Second, using both the original and perturbed instructions, we perform Cross-Region Pair Sampling to derive a comprehensive set of preference pairs, supported by an atomic constraint-based verification. This ensures broad coverage of all relevant regions within the combined response space of the original and perturbed instructions. Figure~\ref{fig2} illustrates an overview of our framework.

\subsection{Cross-Relationship Perturbation}

Given an instruction, we first utilize an advanced LLM to decompose it into its constituent constraints. Our focus is specifically on hard constraints that can be verified through executable code. This decomposition yields a constraint set $\mathcal{C}=\{c_1, ... , c_m\}$, where $m$ denotes the number of target constraints.

% This process produces corresponding sets of perturbed instructions for each relationship. 

% . For every constraint $c_i$, the LLM is guided

 % that induce one of the three specified relationships: containment, partial overlap, or disjoint

Next, for each instruction $x$, we iterate over its constraint set $\mathcal{C}$ and the three predefined relationships to generate perturbations. For example, for the containment relationship, we obtain the set $\mathcal{P}_{x}^{\rm con}=\{\hat{x}_{c_1}, ... ,\hat{x}_{c_m}\}$, where each $\hat{x}_{c_i}$ is a variant of $x$ with constraint $c_i$ perturbed such that the response spaces of $\hat{x}_{c_i}$ and $x$ exhibit a containment relationship. This procedure is applied to all three relationships, resulting in three distinct perturbed instruction sets: $\mathcal{P}_{x}^{\rm con}$, $\mathcal{P}_{x}^{\rm par}$ and $\mathcal{P}_{x}^{\rm dis}$, corresponding to containment, partial overlap and disjoint, respectively. Formally, the Cross-Relationship Perturbation process is described in Algorithm~\ref{alg:algorithm}.

\begin{algorithm}[tb]
    \caption{Cross-Relationship Perturbation}
    \label{alg:algorithm}
    \textbf{Input}: Complex instruction set $\mathcal{X}$ \\
    \textbf{Output}: Perturbed instruction set $\mathcal{P}_{x}^{\rm con},\mathcal{P}_{x}^{\rm par},\mathcal{P}_{x}^{\rm dis}$.\par
    \begin{algorithmic}[1] %[1] enables line numbers
        \STATE Let $\mathcal{P}_{x}^{\rm con}=\mathcal{P}_{x}^{\rm par}=\mathcal{P}_{x}^{\rm dis}=\emptyset$.
        \FOR{$x_j$ in $\mathcal{X}$}
        \STATE $\mathcal{C}_j\leftarrow$ Decompose($x_j$)
        \ENDFOR
        \FOR{$x_j$ in $\mathcal{X}$}
        \FOR{$r$ in 
        \{con, par, dis\}}
        \STATE $\mathcal{P}^{r}_{x_j}=\emptyset$
        \FOR{$c_i$ in $\mathcal{C}_j$}
        \STATE $\hat{c}_i,\hat{x}_{c_i}\leftarrow$ Perturb($c_i,x_j$)
        \STATE $\mathcal{P}^{r}_{x_j}=\mathcal{P}^{r}_{x_j}\cup\hat{x}_{c_i}$
        % \STATE $\hat{\mathcal{C}}^{i}_{j}=(C_j \setminus \{c_i\}) \cup \{\hat{c}_i\}$
        % \STATE $\mathcal{C}^{r}_{x_j}=\mathcal{C}^{r}_{x_j}\cup\hat{\mathcal{C}}^{i}_{j}$
        \ENDFOR
        \STATE $\mathcal{P}^{r}_{x}=\mathcal{P}^{r}_{x}\cup\mathcal{P}^{r}_{x_j}$
        % \STATE $\mathcal{C}^{r}_{x}=\mathcal{C}^{r}_{x}\cup\mathcal{C}^{r}_{x_j}$
        \ENDFOR
        \ENDFOR
        \STATE \textbf{return} $\mathcal{P}_{x}^{\rm con}$, $\mathcal{P}_{x}^{\rm par}$, $\mathcal{P}_{x}^{\rm dis}$
    \end{algorithmic}
\end{algorithm}

\subsection{Cross-Region Pair Sampling}

% For a given relationship and a specific constraint $c_i$, 

% To construct comprehensive preference pairs, 

% Under partial overlap,

The combined response space of the original instruction $x$ and its perturbed variant $\hat{x}_{c_i}$ can be partitioned into two distinct regions, determined by which instructions a given response satisfies: (1) \textbf{Both-Satisfying}, where responses satisfy both the original and the perturbed instruction; (2) \textbf{Single-Satisfying}, where responses satisfy only one of the two instructions.

To ensure preference data captures the full spectrum of a relationship, we perform systematic sampling across these regions. We illustrate this process using the partial overlap relationship as an example. Here, the combined response space divides into three exclusive sub-regions: (1) responses satisfying only the original instruction; (2) responses satisfying only the perturbed instruction; (3) responses satisfying both instructions. This sampling process is formalized as follows:

\begin{equation}
\begin{split}
\mathcal{O}^{\rm original}=\{(x, y^+,y^-)\,|\,y^+ \in \mathcal{S}_1 \,\& \, y^- \in \mathcal{S}_2\} \,\\\cup
\{(x, y^+,y^-)\,|\,y^+ \in \mathcal{S}_3 \,\& \, y^- \in \mathcal{S}_{\rm 2}^{\rm }\}
\end{split}
\end{equation}

\begin{equation}
\begin{split}
\mathcal{O}^{\rm perturbed}=\{(\hat{x}_{c_i},y^+,y^-)\,|\,y^+ \in \mathcal{S}_2 \,\& \, y^- \in \mathcal{S}_1\} \\\,\cup
\{(\hat{x}_{c_i},y^+,y^-)\,|\,y^+ \in \mathcal{S}_3 \,\& \, y^- \in \mathcal{S}_{\rm 1}\}
\end{split}
\end{equation}
where $\mathcal{S}_1$, $\mathcal{S}_2$ and $\mathcal{S}_3$ denote the response sets corresponding to the three sub-regions defined above; $\mathcal{O}^{\rm original}$ and $\mathcal{O}^{\rm perturbed}$ represent the sets of preference pairs for the original instruction and and the perturbed instruction, respectively. 

% adapts based on the origin of the positive response.

Overall, the preference pair composition adopts a contrastive perspective to enhance learning efficiency~\cite{contrastive,iopo}. When a positive response is drawn from an exclusive region (e.g., $\mathcal{S}_1$ or $\mathcal{S}_2$), the corresponding negative is sampled from the opposite exclusive region. This cross-region pairing provides strong contrastive signals, helping the model more effectively distinguish between correct and incorrect constraint satisfaction.

% Region-Across Pair Sampling requires that positive responses be drawn strictly from designated regions within the combined response space. To enforce this requirement, we introduce an atomic constraint-based verification mechanism.

% for the constraint sets associated with both the original and perturbed instructions,

% for each instruction in the original and perturbed instruction sets, we sample $K$ responses, resulting in a response set $Y_j = \{y_1, y_2, \dots, y_K\}$. We then 

 % $x$/$\hat{x}$
\subsubsection{Atomic Constraint-based Verification}
To ensure that positive responses should be drawn strictly from designated regions within the combined response space, we introduce an atomic constraint-based verification mechanism. Specifically, we prompt an LLM to generate a dedicated verification function for each atomic constraint. To guarantee reliability, each function must execute successfully and exceed 80\% accuracy on six LLM-generated test cases; otherwise, it is regenerated. This yields a set of verification functions $V = \{v_1, v_2, \dots, v_m\}$,where each Boolean function $v_i$
 evaluates if a response satisfies constraint $c_i$. Formally, each sampled response $y_k$ is verified against its corresponding constraints by computing:
\begin{equation}
a_k = \prod_{v_i \in {V}_j} v_i(y_k)
\end{equation}

Responses where $a_k=1$ are classified as positive. This approach provides a fine-grained and more strict evaluation of each response against individual constraints.

% not inherently tied to any specific training algorithm. The positive and negative samples it constructs are

 % In this work, we demonstrate this flexibility by utilizing the sampled pairs for SFT, DPO~\cite{dpo}, KTO~\cite{KTO}, and an online variant of DPO~\cite{iopo,bmc,huang-etal-2025-musc}.

 % a variety of training methods, including Supervised Fine-Tuning (SFT) and

\subsection{Training Objectives}
Our framework serves as a general‑purpose generator of preference pairs and is compatible with a range of preference optimization methods. To demonstrate this flexibility, we apply the sampled pairs to DPO~\cite{dpo}, KTO~\cite{KTO}, and an online variant of DPO~\cite{spar}.

\section{Experiments}
\subsection{Setup}
\paragraph{Datasets and Backbones}

 % for training and evaluation
 % In addition to evaluating on the standard IFEval and IFBench datasets,
% Following prior work~\cite{autoif,ifbench}, 

We evaluate our method on two established instruction-following datasets: IFEval~\cite{ifeval} and IFBench~\cite{ifbench}. Following the evaluation settings of prior works~\cite{autoif,ifbench}, for IFEval, we use 433 instructions for training and hold out the remaining 108 as a test set. For IFBench, we randomly sample 2000 instructions from its training split as our training set and use its official test set of 294 instructions for final evaluation. Additionally, we include 200 perturbed instructions generated by our Cross-Relationship Perturbation method as a separate testing set, which we name Perturbed-IF. Furthermore, to assess generalization beyond code-verifiable constraints, we also evaluate our method on the FollowBench dataset~\cite{followbench}. We conduct our experiments using two open-source LLMs as backbones: Qwen2.5-7B-Instruct~\footnote{https://huggingface.co/Qwen/Qwen2.5-7B-Instruct} and Llama-3.1-8B-Instruct~\footnote{https://huggingface.co/meta-llama/Llama-3.1-8B-Instruct}.

% For a testing set $\mathcal{D} = \{(x_j)\}_{j=1}^N$, consisting of $N$ complex instructions where each instruction entails a set of specific constraints $\mathcal{C}_j$. For each input instruction $x_j$, $y_j$ denotes the corresponding response generated by the model. We define two accuracy metrics as follows:

% This metric measures the percentage of satisfied constraints. It

\paragraph{Evaluation Protocol} Consider a test set $\mathcal{D} = \{x_j\}_{j=1}^N$ containing $N$ instructions. Each instruction $x_j$ is associated with a specific set of constraints $\mathcal{C}_j$. For each instruction $x_j$, we denote the model's generated response as $y_i$. We evaluate performance using the following two accuracy metrics:

\clearpage

\begin{table*}[h]
\centering
\footnotesize  
\setlength{\tabcolsep}{1pt} 

\begin{tabularx}{\textwidth}{ 
    l 
    c 
    *{9}{>{\centering\arraybackslash}X} 
}
\toprule
% 表头第一行
\multirow{2}{*}{Method} & \multirow{2}{*}{\makecell{Training \\ size}} &
\multicolumn{2}{c}{IFEval} &
\multicolumn{2}{c}{IFBench} &
\multicolumn{2}{c}{Perturbed-IF} &
\multicolumn{2}{c}{FollowBench} & 
\multirow{2}{*}{Avg.} \\ % 注意这里FollowBench跨两列，凑齐14列

% 横线处理
\cmidrule(lr){3-4}
\cmidrule(lr){5-6}
\cmidrule(lr){7-8}
\cmidrule(lr){9-10}

& &

{$\text{Acc}_{\text{ins}}$} & {$\text{Acc}_{\text{con}}$} & 
{$\text{Acc}_{\text{ins}}$} & {$\text{Acc}_{\text{con}}$} & 
{$\text{Acc}_{\text{ins}}$} & {$\text{Acc}_{\text{con}}$} &
{$\text{Acc}_{\text{ins}}$} & {$\text{Acc}_{\text{con}}$} \\
\midrule
\rowcolor{pink!20}
\multicolumn{11}{c}
{Backbone: Qwen2.5-7B-Instruct}\\
% 数据行
Qwen2.5-7B-Instruct & - & 71.2 & 78.7 & 26.5 & 29.0 & 42.5 & 65.0 & 50.9 & 58.2 & 52.8\\
% \midrule 
% Training strategy: SFT \\
% Rejection Sampling & 2k & 71.3 & 79.9 & 76.9 & 83.3 & \underline{23.5} & \underline{25.1} & 25.9 & 28.4 & 44.5 & 63.5 & 45.1 & 54.8 & 51.9\\
% From Complex to Simple & 1.5k & 64.8 & 74.7 & 67.6 & 77.0 & 17.7 & 20.0 & 20.1 & 24.2 & 38.5 & \underline{69.1} & 40.8 & 52.6 & 47.3\\
% SPAR & 8k & 67.6 & 77.0 & 67.6 & 78.2 & 21.1 & 23.3 & \underline{26.2} & \underline{29.3} & 46.0 & 64.6 & \textbf{50.7} & \textbf{58.6} & 50.9\\
% UltraIF & 10k & 58.1 & 67.8 & 60.2 & 69.4 & 20.7 & 22.7 & 22.8 & 25.3 & 41.5 & 60.4 & 37.3 & 47.0 & 44.4\\
% AutoIF & 10k & 63.9 & 71.8 & 67.6 & 75.3 & 17.0 & 17.9 & 22.4 & 24.5 & 41.0 & 62.4 & \underline{45.7} & \underline{55.6} & 47.1\\
% Our(IFEval) & 12k & \underline{75.9} & \underline{82.2} & \textbf{78.7} & \underline{83.9} & 22.1 & 22.4 & 24.5 & 25.7 & \underline{47.0} & 67.5 & 45.1 & 54.3 & \underline{52.4}\\
% Our(IFBench) & 12k & \textbf{76.6} & \textbf{83.3} & \underline{78.2} & \textbf{85.1} & \textbf{27.2} & \textbf{29.0} & \textbf{29.1} & \textbf{31.3} & \textbf{52.0} & \textbf{70.1} & 45.4 & 54.9 & \textbf{55.2}\\
\midrule 
\rowcolor{gray!20} \multicolumn{11}{c}{Training strategy: DPO}\\
Rejection Sampling & 12k & 75.7 & 81.0 & 24.8 & 26.3 & 51.0 & 66.6 & 46.2 & 55.6 & 53.4\\
From Complex to Simple & 12k & 74.1 & 81.6 & 27.6 & 29.9 & 45.5 & 64.8 & 51.6 & 58.5 & 54.2\\
AutoIF & 12k & 74.1 & 81.0 & 31.3 & \underline{35.2} & 43.5 & 66.8 & 52.5 & 59.6 & 55.5\\
$\text{Ours}_\text{IFEval} $& 12k  & \textbf{90.7} & \textbf{94.3} & \underline{32.0} & 34.6 & \textbf{56.5} & \textbf{73.5} & \underline{55.4} & \underline{61.8} & \textbf{62.4}\\
$\text{Ours}_\text{IFBench} $ & 12k & \underline{82.2} & \underline{86.6} & \textbf{34.9} & \textbf{36.2} & \underline{54.0} & \underline{72.8} & \textbf{57.5} & \textbf{63.2} & \underline{60.9}\\
\midrule 
\rowcolor{gray!20} \multicolumn{11}{c}{Training strategy: KTO}\\
Rejection Sampling & 12k & 70.4 & 75.9 & 23.5 & 25.7 & 44.5 & 62.4 & 45.5 & 55.1 & 50.4 \\
From Complex to Simple & 12k & 73.1 & 80.5 & 28.2 & 30.7 & 43.5 & 65.0 & 49.4 & 56.6 & 53.4 \\
AutoIF & 12k & 72.2 & 80.5 & 29.9 & 31.6 & 42.5 & 65.0 & \underline{52.5} & \underline{59.8} & 54.3\\
$\text{Ours}_\text{IFEval} $& 12k & \textbf{82.4} & \textbf{87.4} & \textbf{30.6} & \underline{31.9} & \textbf{52.0} & \textbf{71.5} & \textbf{54.7} & \textbf{61.8} & \textbf{59.0}\\
$\text{Ours}_\text{IFBench} $ & 12k & \underline{78.7} & \underline{83.6}  & \underline{30.4} & \textbf{32.6} & \underline{49.5} & \underline{68.8} & 51.5 & 58.1 & \underline{56.7}\\
\midrule 
\rowcolor{gray!20} \multicolumn{11}{c}{Training strategy: Online DPO}\\
% \midrule 

UltraIF & 12k & 67.1 & 73.9 & 24.4 & 26.2 & 43.5 & 60.8 & 45.1 & 54.9 & 49.5 \\
SPAR & 12k & 72.2 & 80.3 & \textbf{33.7} & \textbf{36.0}  & 42.0 & 65.3 & 29.6 & 40.1 & 49.9\\
$\text{Ours}_\text{IFEval} $& 12k & \textbf{76.9} & \textbf{82.9} & \underline{33.3} & \underline{34.6} & \textbf{49.5} & \textbf{70.1} & \textbf{53.2} & \underline{59.5} & \textbf{57.5}\\
$\text{Ours}_\text{IFBench} $ & 12k & \underline{75.0} & \underline{81.8} & 28.5 & 30.1  & \underline{48.0} & \underline{68.6} & \underline{53.1} & \textbf{59.6} & \underline{55.6}\\
\midrule

\rowcolor{pink!20}
\multicolumn{11}{c} 
{Backbone: Llama3.1-8B-Instruct}\\
Llama3.1-8B-Instruct & - & 80.6 & 85.8 & 22.5 & 24.4 & 49.0 & 70.0 & 51.5 & 58.4 & 55.3\\
\midrule 
\rowcolor{gray!20} \multicolumn{11}{c}{Training strategy: DPO}\\
Rejection Sampling & 12k & 81.5 & 87.6 & 23.8 & 26.4 & 57.5 & 75.7 & 46.8 & 53.2 & 56.6\\
From Complex to Simple & 12k & 76.9 & 83.3 & 33.7 & 35.8 & 53.0 & 73.9 & 50.8 & 58.1 & 58.2\\
AutoIF & 12k & 84.3 & 89.7 & \underline{35.7} & \textbf{38.8} & 55.0 & 75.2 & 49.3 & 57.8 & 60.7\\
$\text{Ours}_\text{IFEval} $& 12k & \underline{88.0} & \underline{92.5} & \textbf{36.4} & \underline{37.9} & \underline{65.0} & \underline{81.6} & \underline{55.1} & \underline{60.4} & \underline{64.6}\\
$\text{Ours}_\text{IFBench} $ & 12k & \textbf{88.7} & \textbf{92.7} & 34.4 & 37.5 & \textbf{66.5} & \textbf{82.1} & \textbf{57.2} & \textbf{62.2} & \textbf{65.2}\\
\midrule 
\rowcolor{gray!20} \multicolumn{11}{c}
{Training strategy: KTO}\\
Rejection Sampling & 12k & 79.9 & 85.3 & 25.9 & 28.1 & 54.0 & 75.4 & 50.9 & 57.3 & 57.1\\
From Complex to Simple & 12k & 78.7 & 85.6 & 27.6 & 29.6 & 49.5 & 71.2 & 51.3 & 58.6 & 56.5\\
AutoIF & 12k & \underline{80.6} & 86.8 & \underline{31.7} & \textbf{34.9} & \underline{61.5} & \underline{78.7} & 54.6 & \textbf{62.0} & 61.4\\
$\text{Ours}_\text{IFEval} $& 12k & \textbf{86.6} & \textbf{91.5} & \textbf{32.6} & \underline{34.6} & \textbf{64.5} & \textbf{80.8} & \underline{54.7} & 61.4 & \textbf{63.3}\\
$\text{Ours}_\text{IFBench} $ & 12k & \textbf{86.6} & \underline{91.4} & 31.1 & 34.1 & 56.5 & 77.2 & \textbf{56.2} & \underline{61.6} & \underline{61.8}\\
\midrule 
\rowcolor{gray!20} \multicolumn{11}{c}
{Training strategy: Online DPO}\\
UltraIF & 12k & 83.8 & 88.2 & 26.0 & \underline{28.3} & \textbf{54.5} & \underline{74.6} & 47.8 & 55.0 & 57.3\\
SPAR & 12k & 79.9 & 86.8 & \textbf{27.7} & \textbf{29.7} & 52.0 & 73.7 & 43.7 & 52.3 & 55.7\\
$\text{Ours}_\text{IFEval} $& 12k & \textbf{85.8} & \textbf{90.5} & \underline{27.6} & \underline{28.3} & \underline{54.0} & \textbf{77.0} & \underline{52.5} & \textbf{58.7} & \textbf{59.3}\\
$\text{Ours}_\text{IFBench} $ & 12k & \underline{84.7} & \underline{89.2} & 26.0 & 27.3 &  52.0 & 74.3 & \textbf{52.6} & \underline{58.3} & \underline{58.1}\\
\midrule 
\rowcolor{gray!20} \multicolumn{11}{c}{Recent baselines trained with RLVR on R1-Distill-Qwen-7B and TULU-3-8B} \\
VerIF (R1-Distill-Qwen-7B) & 22k & 73.6 & 82.9 & 18.8 & 21.0 & 26.5 & 47.8 & 47.6 & 55.1 & 46.7\\
VerIF (TULU-3-8B) & 22k & 85.0 & 89.8  & 23.6 & 26.3 & 55.5 & 77.7 & 53.4 & 60.4 & 59.0\\
\bottomrule
\end{tabularx}
\caption{Performance comparison of our method against prior baselines across the four test sets. For each backbone and training strategy, the best and second-best accuracy scores are highlighted in \textbf{bold} and \underline{underlined}, respectively. $\text{Ours}_\text{IFEval}$ and $\text{Ours}_\text{IFBench}$ indicate models trained on the IFEval and IFBench training sets, respectively.}
\label{tab:1}

\end{table*}

\begin{table*}[t]
\centering
\footnotesize
\setlength{\tabcolsep}{1pt}

\begin{tabularx}{\textwidth}{
    l
    *{10}{>{\centering\arraybackslash}X}
}
\toprule

% 表头第一行
\multirow{2}{*}{Model} &
\multicolumn{2}{c}{Original} &
\multicolumn{2}{c}{Containment-1} &
\multicolumn{2}{c}{Containment-2} &
\multicolumn{2}{c}{Partial Overlap} &
\multicolumn{2}{c}{Disjoint} \\

% 分组横线
\cmidrule(lr){2-3}
\cmidrule(lr){4-5}
\cmidrule(lr){6-7}
\cmidrule(lr){8-9}
\cmidrule(lr){10-11}

% 表头第二行
& $\text{Acc}_{\text{ins}}$ & $\text{Acc}_{\text{con}}$ & $\text{Acc}_{\text{ins}}$ & $\text{Acc}_{\text{con}}$  & $\text{Acc}_{\text{ins}}$ & $\text{Acc}_{\text{con}}$ & $\text{Acc}_{\text{ins}}$ & $\text{Acc}_{\text{con}}$ & $\text{Acc}_{\text{ins}}$ & $\text{Acc}_{\text{con}}$ \\

\midrule

% 数据行
Qwen2.5-7B-Instruct & 71.2 & 78.7 & 53.7 & 71.0 & 58.8 & 73.2 & 50.0 & 67.9 & 51.2 & 70.1\\
Rejection Sampling & 75.7 & 81.0 & 57.5 & 75.0 & 60.0 & 75.6 & 55.0 & 72.8 & 62.5 & 73.7\\
From Complex to Simple & 84.1 & 81.6 & 62.5 & 76.7 & 60.0 & 72.6 & 50.0 & 69.1 & 53.7 & 70.1\\
AutoIF & 84.1 & 81.0 & 51.2 & 72.7 & 57.5 & 73.2 & 50.0 & 70.4 & 51.2 & 71.9\\
\hline
Ours & 90.7 & 94.3 & 68.8 & 80.7 & 67.5 & 80.4 & 62.5 & 79.6 & 68.8 & 80.2\\

\hline
\end{tabularx}

\caption{Performance on original instructions and four types of perturbed instructions. Containment-1 and Containment-2 represent two subtypes of containment relationships, namely containing and contained.}
\label{tab:error-ana}
\end{table*}
\clearpage

\begin{itemize}
\item \textbf{{Constraint-level Accuracy ($\text{Acc}_{\text{con}}$)}} This metric computes the proportion of satisfied constraints relative to the total number of constraints across all instructions:

\begin{equation}
    \text{Acc}_{\text{con}} = \frac{\sum_{j=1}^N \sum_{c \in \mathcal{C}_j} \mathbb{V}(y_j, c)}{\sum_{j=1}^N |\mathcal{C}_j|}
\end{equation}
where $\mathbb{V}(\cdot,\cdot)$ denotes the verification function.

\item \textbf{{Instruction-level Accuracy ($\text{Acc}_{\text{ins}}$)}} This metric computes the proportion of instructions for which all constraints are satisfied across the test set:
\begin{equation}
    \text{Acc}_{\text{ins}} = \frac{1}{N} \sum_{j=1}^N \left( \prod_{c \in \mathcal{C}_j} \mathbb{V}(y_j, c) \right)
\end{equation}
where $\prod$ is a logical AND operation, equaling 1 only if the response satisfies all constraints for instruction $x_j$.
\end{itemize}

\paragraph{Baselines} 
We categorize the preference optimization baselines into two types: online and offline. Offline baselines include \textbf{Rejection Sampling}~\cite{reject1}, which constructs preference pairs by sampling and verifying multiple responses per instruction; \textbf{From Complex to Simple}~\cite{fcs}, which forms preference pairs through teacher correction of student outputs; and \textbf{Autoif}~\cite{autoif}, which employs LLMs to automatically generate instructions and verification code. We compare these baselines with our method across both DPO and KTO training strategies.

Online baselines include \textbf{UltraIF}~\cite{ultraif}, which trains an instruction composer to iteratively synthesize constrained complex instructions, followed by sampling and evaluation for online DPO training; and \textbf{SPAR}~\cite{spar}, which constructs preference data for online DPO by performing tree-search–based self-corrections on instruction-violating responses. We compare these baselines with our method using the online DPO framework adopted in SPAR.

% For a fair comparison with these baselines, we evaluate them using the same backbone models: Qwen2.5-7B-Instruct and Llama-3.1-8B-Instruct.

We also compare with recent baselines that employ different training strategies. For instance, 
\textbf{VerIF} \cite{verif} combines rule-based code verification with LLM-based validation to perform RLVR.

\paragraph{Implementation Details}
We employ DeepSeek v3.1 for instruction decomposition, perturbation, and the generation of verification functions. For offline training, DeepSeek v3.1 is also used for response sampling. In online training, responses are sampled from the optimized model at each iteration. All experiments are run on a single A100 GPU using the OpenRLHF framework~\cite{openrlhf}.

\subsection{Main Results}

The performance comparison is presented in Table~\ref{tab:1}. We highlight three key observations:

\paragraph{Performance on Offline Training}
Our framework significantly enhances instruction-following performance across both DPO and KTO training methods, as well as all four test sets. For instance, under the DPO training strategy with the Qwen2.5-7B-Instruct backbone, our models trained on either IFEval or IFBench consistently outperform prior baselines. Specifically, Our$_{\rm IFEval}$ surpasses AutoIF by 6.9\% in average accuracy across the four datasets. Similar performance gains are observed with the KTO training strategy. These results demonstrate the superiority and generalization of our method.

\paragraph{Performance on Online Training}  Under online training, without supervision from advanced LLMs, the performance of prior baselines is limited on the Qwen2.5-7B-Instruct backbone. Both UltraIF and SPAR exhibit lower average accuracy compared to the pretrained Qwen2.5-7B-Instruct. Nevertheless, our method maintains significant performance improvement in this setting. This can be contributed to our method's diverse sampling and atomic verification, which enable the acquisition of higher-quality preference pairs.

\begin{figure*}[t]
  \includegraphics[width=0.48\textwidth]{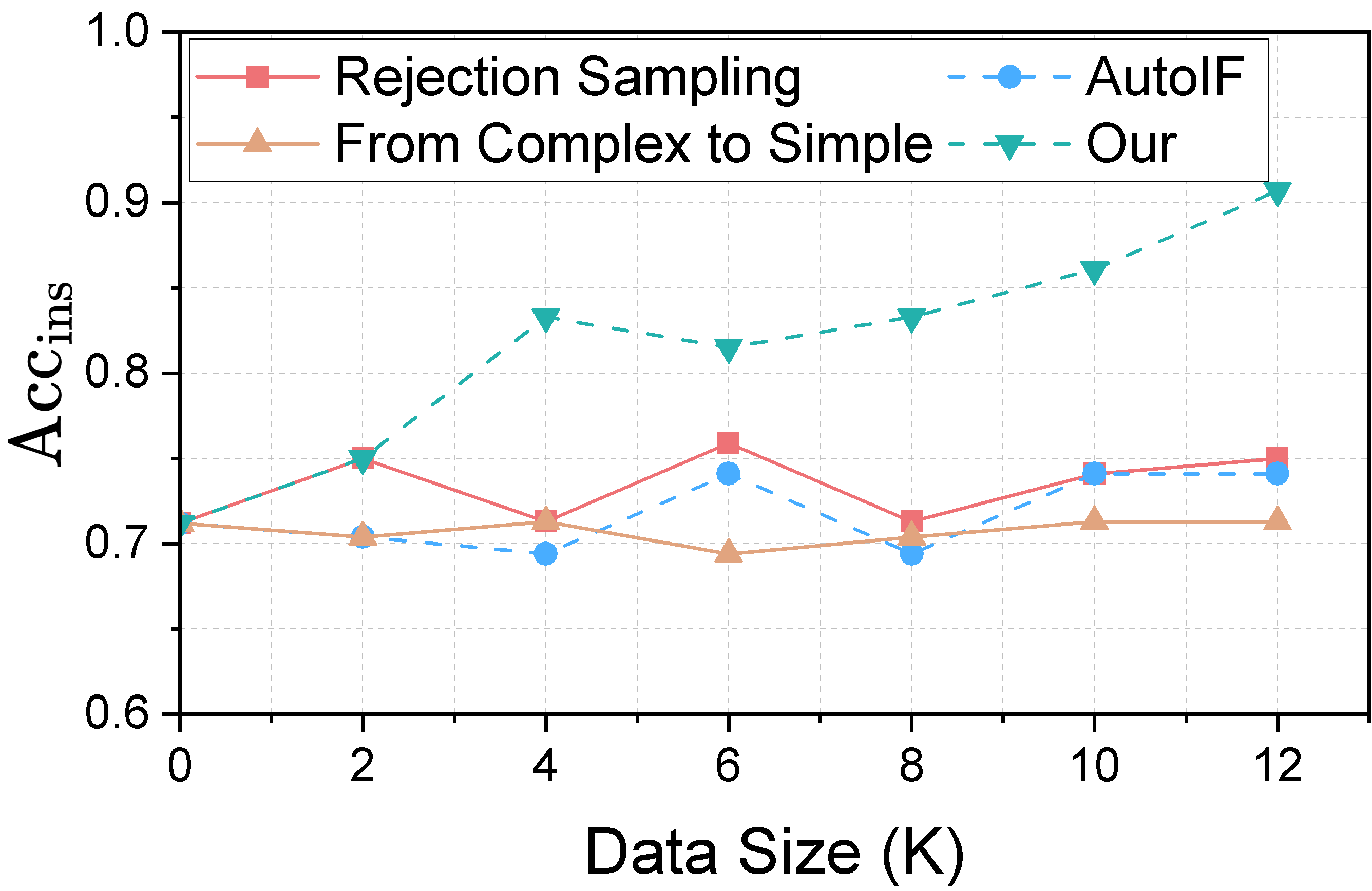} \hfill
  \includegraphics[width=0.48\textwidth]{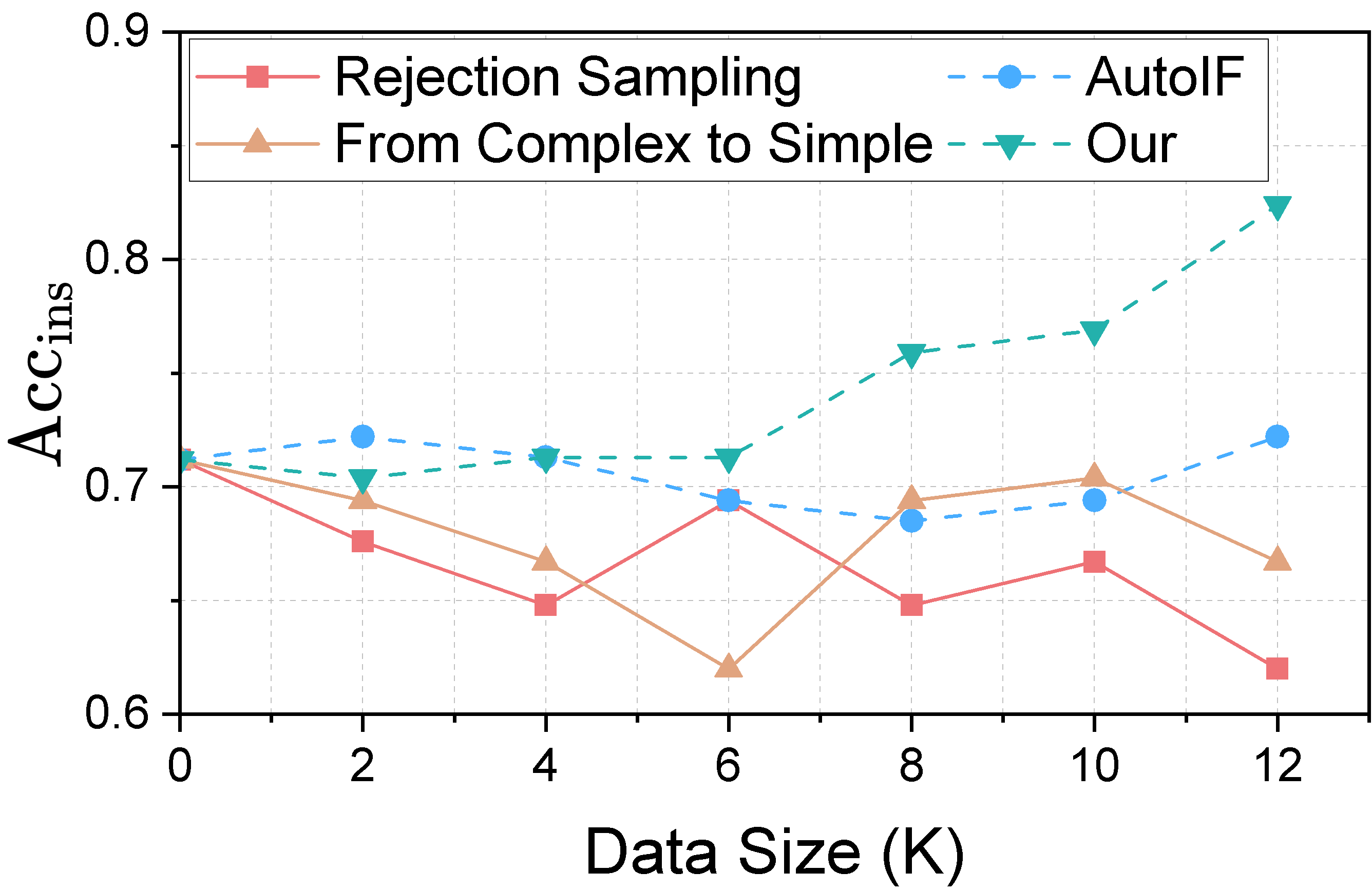}
  \caption{Performance scaling with training data size for DPO (left) and KTO (right) across different methods.}
  \label{scale}
\end{figure*}

\paragraph{Performance across Different Backbones} Our method demonstrates performance gains across both LLM backbones,  further validating the generalization of our approach. We also conducted training on Qwen2.5-14B-Instruct~\footnote{https://huggingface.co/Qwen/Qwen2.5-14B-Instruct} and Qwen3-8B~\footnote{https://huggingface.co/Qwen/Qwen3-8B}, and the results are provided in the Appendix.

\begin{table}[t]
\centering
\footnotesize  
\setlength{\tabcolsep}{1pt} 

\begin{tabularx}{0.7\columnwidth}{
    l
    *{2}{>{\centering\arraybackslash}X}
}
\toprule
Method & $\text{Acc}_{\text{ins}}$ & $\text{Acc}_{\text{con}}$ \\
\midrule

% 数据行
$\text{Ours} $  & 90.7 & 94.3 \\
% \hline
w/o Containment-1 & 85.9 & 90.7  \\
w/o Containment-2 & 84.7 & 88.8 \\
w/o Partial Overlap & 84.3 & 88.8 \\
w/o Disjoint & 82.9 & 87.2\\
\bottomrule
\end{tabularx}
\caption{Ablated results on different relationship types.}
\label{abla}
\end{table}

\begin{table}[t]
\centering
% 1. 缩小字号，14列通常需要 scriptsize
\footnotesize  
% 2. 减小列与列之间的间距 (默认是6pt，太宽了，改为 2pt 或 1pt)
\setlength{\tabcolsep}{1pt} 

% 3. 定义列格式
\begin{tabularx}{0.7\columnwidth}{
    l
    *{2}{>{\centering\arraybackslash}X}
}
\toprule
Method & $\text{Acc}_{\text{ins}}$ & $\text{Acc}_{\text{con}}$ \\
\midrule

% 数据行
$\text{Ours} $  & 81.3 & 85.6 \\
% \hline
w/o Pair1 & 77.1 & 82.5  \\
w/o Pair2 & 77.3 & 83.1 \\
w/o Pair3 & 78.5 & 83.5 \\
w/o Pair4 & 79.2 & 84.2\\
% Only-Disjoint & 82.9 & 87.2 & 85.9 & 89.2 \\
\bottomrule
\end{tabularx}
\caption{Ablated results on cross-region pair sampling.}
\label{abla-pair}
\end{table}

\begin{table}[t]
\centering
\footnotesize  

\begin{tabularx}{0.9\columnwidth}{
    l
    *{4}{>{\centering\arraybackslash}X}
    *{2}{>{\centering\arraybackslash}X}
}
\toprule
% 表头第一行
Method & TP & FP & TF & NF & $\text{Acc}_{\text{ins}}$ & $\text{Acc}_{\text{con}}$ \\
\midrule

% 数据行
ACV (ours) & 66 & 0 & 27 & 15 & 85.4 & 89.2 \\

ILV  & 47 & 0 & 27 & 34 & 84.0 & 88.7 \\
LAJ  & 52 & 4 & 23 & 29 & 80.8 & 85.8\\
\bottomrule
\end{tabularx}
\caption{Performance comparison for different verification methods.}
\label{abla-judge}
\end{table}

\subsection{Further Analysis}

Due to limited computational resources, we conduct further analysis only on Qwen2.5-7B-Instruct using offline training strategies, with IFEval as training set.

 % and evaluated on the IFEval test set

\subsubsection{Performance on Different Data Sizes}
To evaluate the quality of preference pairs generated by different sampling methods, we compare our method with prior baselines under both DPO and KTO as the training data size grows (Figure~\ref{scale}). As training data size increases, our method demonstrates a clear upward performance trend, while other methods show limited gains. These results indicate that our improvements stem not only from increased data quantity but also from higher data quality and more informative preference signals.

\subsubsection{Performance on Different Relationship Types}
A model with strong instruction‑following ability should adhere consistently to a constraint across its variants. We compare our method with baselines—Rejection Sampling, From Complex to Simple, and AutoIF—on the original instructions and four perturbed types (Table~\ref{tab:error-ana}).

Baselines improve performance on the original instructions but show limited gains on perturbed variants. For instance, From Complex to Simple improves only on ``Containment‑1'', while other perturbed types remain close to the pre‑trained model’s performance. Without explicitly accounting for response‑space relationships, these methods struggle to capture comprehensive constraint variations. Rejection Sampling improves more consistently across perturbation types, likely because its random sampling may incidentally cover such variations. In contrast, our method achieves significant improvement on both the original instructions and all perturbed types, demonstrating strong generalization. This underscores the value of training on diverse perturbed instructions for boosting instruction‑following capability.

% \paragraph{General abilities are not declined.}

\subsubsection{Ablation Study}
\paragraph{Relationship Types}

We evaluate the contribution of each relationship type by removing individual relationships from the perturbation process. Table~\ref{abla} presents the results on the IFEval dataset. We use w/o to denote the variant without a specific type. Compared to using the full set of relationships, each removal results in a performance degradation, indicating that including all relationship types during sampling is crucial. Specifically, the ``Disjoint'' relationship type plays the most significant role that provides strong contrastive signals to enhance training.

\paragraph{Cross-Region Pair Sampling}
Taking the ``partial overlap'' relationship as a case study, we assess the contribution of preference pairs drawn from different sampling regions. Specifically, we remove each of the four pair types in turn and report the results in Table~\ref{abla-pair}. Performance degrades when any single pair type is excluded, demonstrating the importance of constructing preference pairs across these regions to ensure sampling diversity.

\paragraph{Verification Methods} To evaluate the effectiveness of our Atomic Constraint-based Verification (ACV), we compare it with two alternative verification strategies: (1) using an LLM as a judge to directly assess whether a response follows the instruction (LLM-as-Judge, LAJ), and (2) generating a single verification function for an instruction (Instruction-level Verification, ILV).

Specifically, we first collect an LLM’s responses to the IFEval test instructions and evaluate them using all three methods. Following the official IFEval validation protocol, we compute the numbers of true positives (TP), false positives (FP), true negatives (TN), and false negatives (FN) for each method, as reported in Table \ref{abla-judge}. ACV achieves the lowest misjudgment rate and yields zero false positives, indicating its superior effectiveness in evaluating a response's adherence to an instruction.

We then apply ILV and LAJ to the IFEval training set to construct preference data and train models on the resulting datasets. ACV achieves the highest performance, further demonstrating that our method more effectively ensure the quality of preference pairs.

\section{Conclusion}
We propose Cross-Relational Preference Learning (CRPL), a novel framework to enhance LLM instruction following. CRPL explicitly models fine‑grained response‑space relationships between different instructions, enabling the generation of preference data that captures a wide spectrum of constraint variations. The processes are rigorously guaranteed by an atomic constraint‑based verification mechanism, which ensures the quality of the sampled pairs. Through experiments across diverse training methods and LLM backbones, our framework demonstrates strong performance on the instruction-following benchmarks.

\clearpage
\section*{Acknowledgments}

This research was partially supported by the Key Research and Development Project in Shaanxi Province No. 2024PT-ZCK-89, the National Natural Science Foundation of China No. 62406242, 62476215, 62302380, 62037001, 62137002 and 62192781, Project of China Knowledge Centre for Engineering Science and Technology.
\bibliographystyle{named}
\bibliography{main}

\clearpage

\appendix

\section{Experimental Setup}
\subsection{Datasets}
\label{Datasets} 
\textbf{IFEval} \cite{ifeval} is a key benchmark for testing how well LLMs follow specific, verifiable instructions. It constructs approximately 540 instructions containing 25 types of verifiable constraints, such as format (JSON, lists), length, capitalization, and content constraints.We use strict accuracy metrics at prompt and instruction levels, corresponding to our $\text{Acc}_{\text{ins}}$ and $\text{Acc}_{\text{con}}$ in this paper.

\textbf{IFBench} \cite{ifbench} is a newer and more challenging benchmark for precise instruction following, containing 58 types of constraints and corresponding verification functions. Similarly, we report strict accuracy metrics at prompt and instruction levels as $\text{Acc}_{\text{ins}}$ and $\text{Acc}_{\text{con}}$.

\textbf{Followbench} \cite{followbench} a multi-level fine-grained constraints following benchmark designed to systemically and precisely evaluate the instruction-following capability of LLMs. It covers five types of constraints (Content, Situation, Style, Format, and Example) and employs a multi-level mechanism that progressively increases constraint difficulty. Importantly, FollowBench relies on strong LLMs as evaluators, Importantly, FollowBench relies on strong LLMs as evaluators, and is therefore used to assess the generalization of our method on instructions for which code-based verification is not applicable. In our experiment, we use GPT-4o-mini as the evaluator.

\subsection{Baselines}
\label{Baselines} 
\textbf{Rejection sampling} \cite{reject1} generates multiple responses for a given prompt and verifies them, where correct and incorrect responses are paired to construct preference data for training. For the offline methods DPO and KTO, to ensure a fair comparison with our approach, we also use DeepSeek v3.1 to sample positive responses, while negative responses are sampled from the model being trained. For online DPO, both positive and negative responses are sampled from the model from the previous training iteration.

\textbf{From Complex to Simple} \cite{fcs} obtains positive and negative pairs via teacher correction of student outputs. We reproduced From Complex to Simple using the preference data provided by the official release~\footnote{https://github.com/meowpass/FollowComplexInstruction}.

\textbf{AutoIF} uses LLMs to generate instructions and code to verify instruction responses, and constructs preference data via rejection sampling based on code-based verification. We reproduced AutoIF using its official open-source code~\footnote{https://github.com/QwenLM/AutoIF}.

\textbf{UltraIF} \cite{ultraif} designs a self-alignment framework that iteratively synthesizes complex instructions and performs sampling and evaluation to construct preference pairs for online DPO training. We reproduced UltraIF  using its official open-source dataset and code~\footnote{https://github.com/kkk-an/UltraIF}.

\textbf{Spar} \cite{spar} also proposes a self-alignment approach that constructs preference data for online DPO training by performing tree-search–based self-corrections on erroneous responses. We reproduced Spar on the Qwen2.5-7B-Instruct and Llama3.1-8B-Instruct backbones using its official open-source  code~\footnote{https://github.com/thu-coai/SPaR}.

\textbf{VerIF} \cite{verif} combines rule-based code verification  with LLM-based validation to perform RLVR. We directly use their released model checkpoints for evaluation and comparison~\footnote{https://github.com/THU-KEG/VerIF?tab=readme-ov-file}.
\subsection{Training Configuration}
\label{subsec:training_config}

\begin{table}[t]
\centering
\footnotesize  

\setlength{\tabcolsep}{1pt} 

% 3. 定义列格式s
\begin{tabularx}{0.7\columnwidth}{
    l
    *{2}{>{\centering\arraybackslash}X}
}
\toprule
Parameter & DPO & KTO \\
\midrule

% 数据行
Learning Rate & 1e-5 & 1e-5 \\
% \hline
Train Batch Size & 32 & 32  \\
Micro Train Batch Size & 8 & 8 \\
Max Sequence Length & 5102 & 5102 \\
Beta & 0.1 & 0.1 \\
Max Epochs & 2 & 3\\
Lora Rank & 16 & 16\\
Lora Alpha & 32 & 32\\
Lora Dropout & 0.1 & 0.1\\
% Only-Disjoint & 82.9 & 87.2 & 85.9 & 89.2 \\
\bottomrule
\end{tabularx}
\caption{Ablated results on different relationship types.}
\label{training}
\end{table}

For each preference learning method, we use LoRA adapters \cite{hu2022lora} for efficient training. The training hyperparameters are provided in Table \ref{training}. The settings for online DPO are consistent with those for offline DPO.

\subsection{Generation Configuration}
During evaluation, for each test set and each backbone LLM, we set the temperature to 0.7 and the maximum generation length to 4096. For each test instance, we sample four responses and report the average score across these samples as the final result for the test set.
\section{Additional Experimental Results}
\label{Additional} 
To further validate the effectiveness of our CRPL framework on different backbones, we conducted DPO and KTO training on Qwen2.5-14B-Instruction and Qwen3-8B. The results are provided in Table \ref{tab:backbone}. It can be seen that applying our framework on larger LLMs can also achieve significant improvements on the test set.
\begin{table}[t]
\centering
\footnotesize  
\setlength{\tabcolsep}{2pt} 

% 3. 定义列格式
\begin{tabularx}{\columnwidth}{
    l
    c
    *{8}{>{\centering\arraybackslash}X}
}
\toprule
% 表头第一行
\multirow{2}{*}{Model} &
\multirow{2}{*}{Data} &
\multicolumn{2}{c}{IFEval} & \multicolumn{2}{c}{IFBench} \\
\cmidrule(lr){3-4}
\cmidrule(lr){5-6}
& &
$\text{Acc}_{\text{ins}}$& $\text{Acc}_{\text{con}}$ & $\text{Acc}_{\text{ins}}$ & $\text{Acc}_{\text{con}}$\\
\midrule

% 数据行
\multirow{3}{*}{\centering Qwen2.5-14B-Instruct} & - & 85.0 & 89.1 & 33.2 & 35.4  \\

& $\text{Ours}_\text{IFEval} $& 88.2 & 91.5 & 36.1 & 38.1 \\
& $\text{Ours}_\text{IFBench} $ & 88.2 & 92.1 & 41.6 & 44.3 \\
\midrule
\multirow{3}{*}{\centering Qwen3-8B} & - & 89.4 & 93.4 & 25.6 & 28.0 \\

& $\text{Ours}_\text{IFEval} $& 89.1 & 93.1 & 29.6 & 31.3 \\
& $\text{Ours}_\text{IFBench} $ & 89.8 & 93.5 & 28.5 & 31.7 \\
\bottomrule
\end{tabularx}
\caption{Performance of Our Method across Different Backbone Models.
}
\label{tab:backbone}
\end{table}

\section{Prompts of CRPL}
\begin{figure}[h]
  \includegraphics[width=\columnwidth]{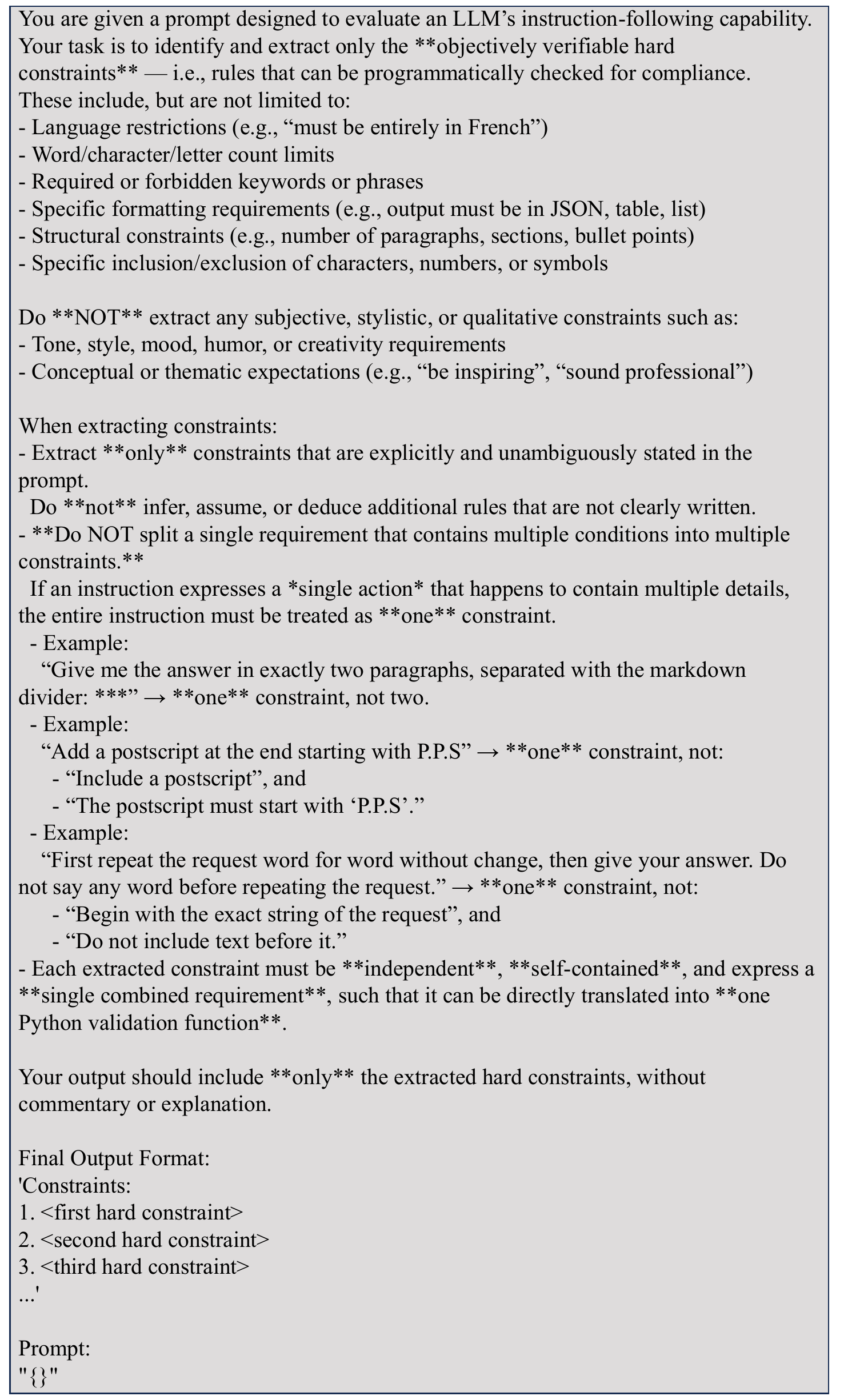}
  \caption{Prompt for instruction decomposition}
  \label{comp}
\end{figure}

\begin{figure}[htbp]
  \includegraphics[width=\columnwidth]{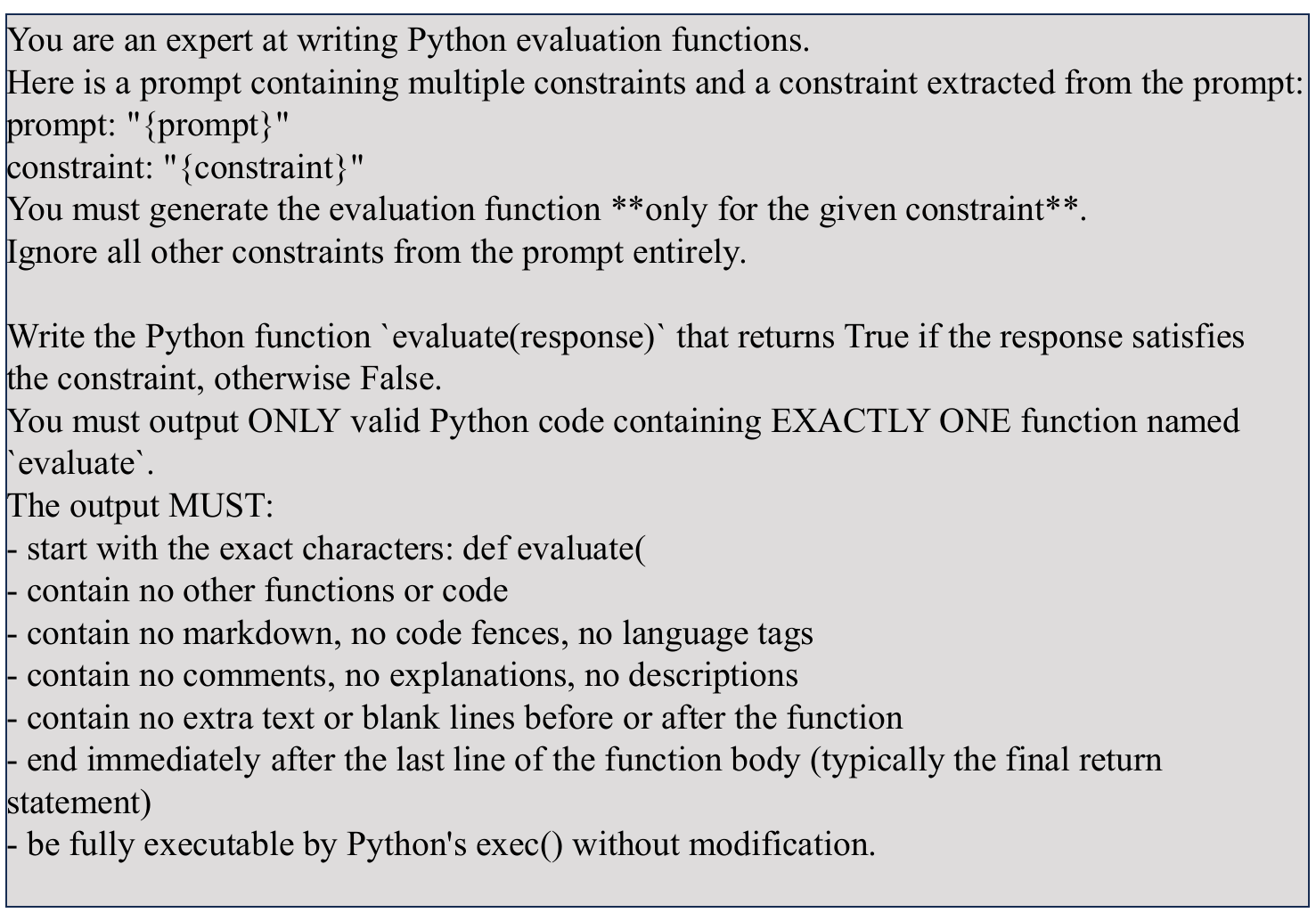}
  \caption{Prompt for generating evaluation functions for atomic constraints.}
  \label{judge}
\end{figure}

\begin{figure}[t]
  \includegraphics[width=\columnwidth]{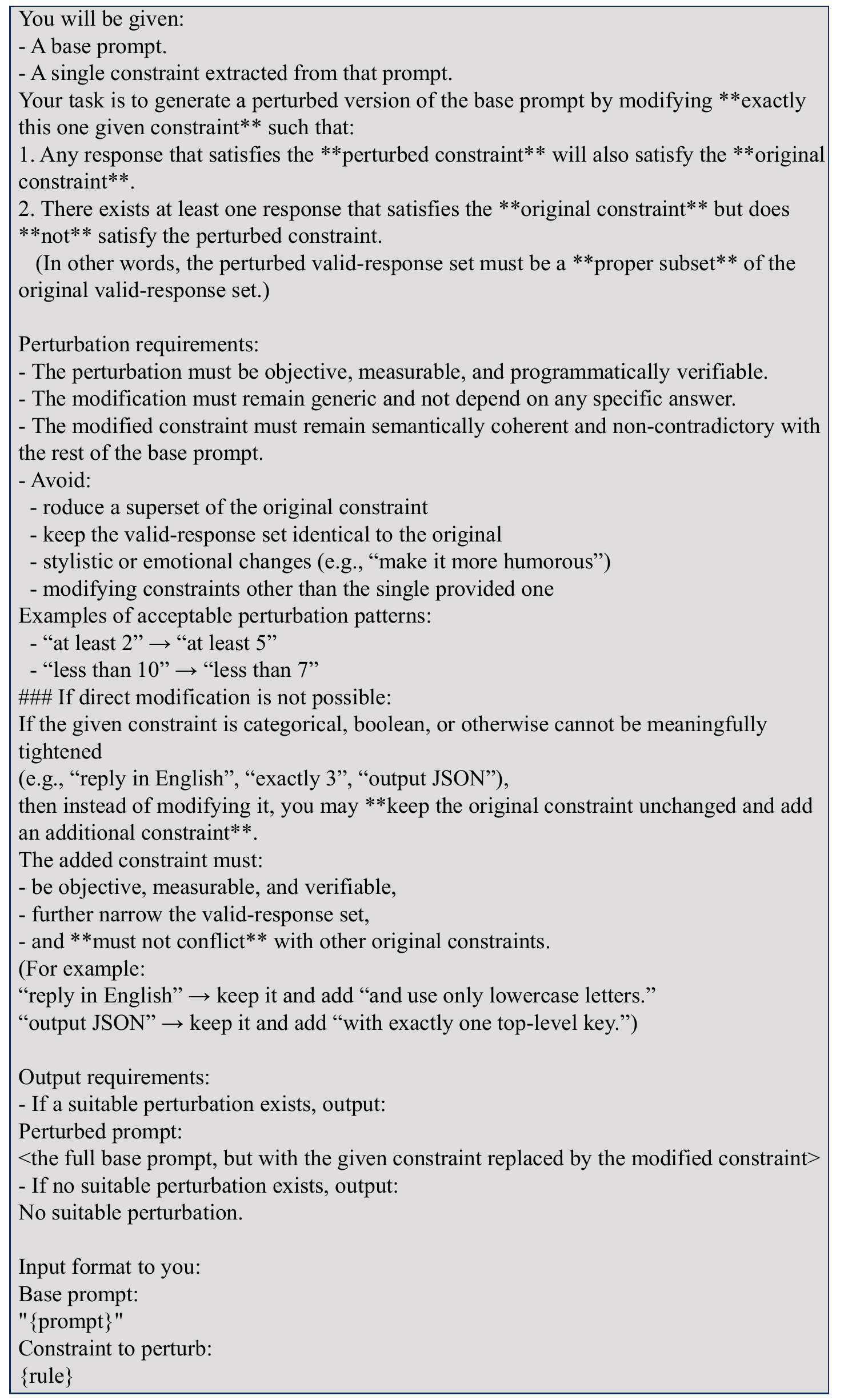}
  \caption{Prompt for the first type of containment relationship perturbation.
The valid response space of the original instruction contains that of the perturbed instruction.}
  \label{pert1}
\end{figure}

\begin{figure}[t]
  \includegraphics[width=\columnwidth]{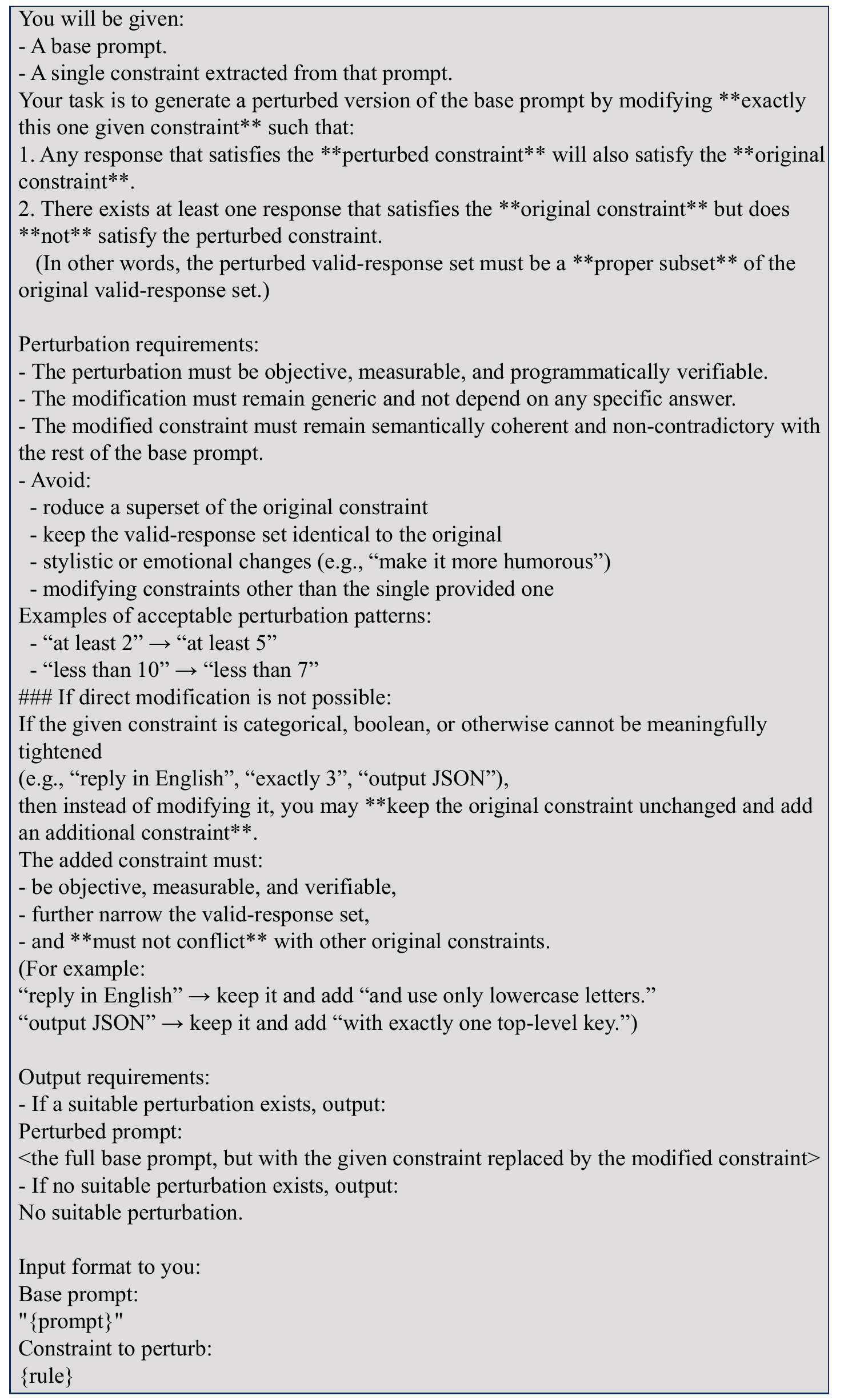}
  \caption{Prompt for the second type of containment relationship perturbation.
The valid response space of the perturbed instruction contains that of the original instruction.}
  \label{pert2}
  
\end{figure}

\begin{figure}[t]
  \includegraphics[width=\columnwidth]{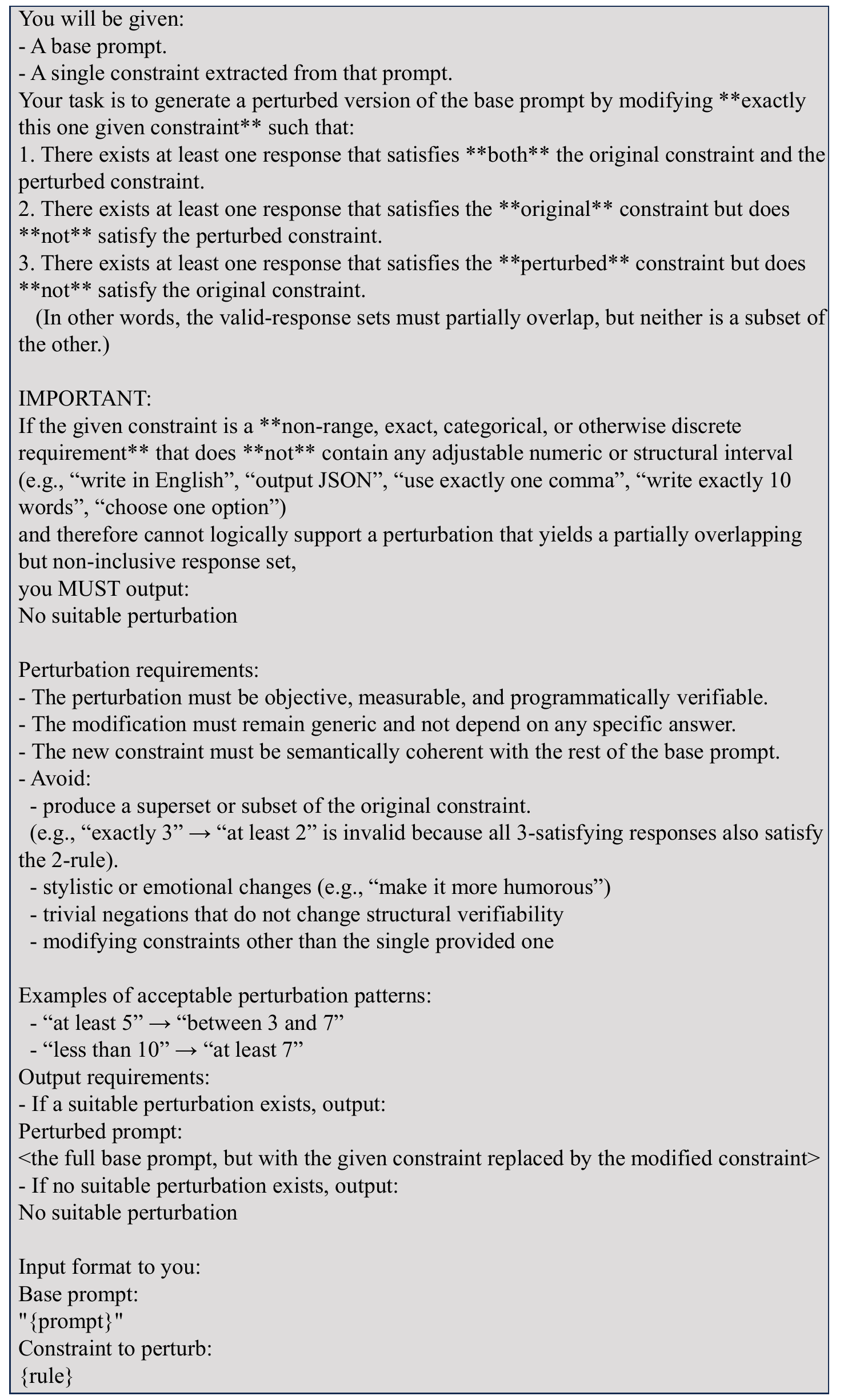}
  \caption{Prompt for the  partial overlap relationship perturbation.}
  \label{pert3}
  
\end{figure}

\begin{figure}[t]
  \includegraphics[width=\columnwidth]{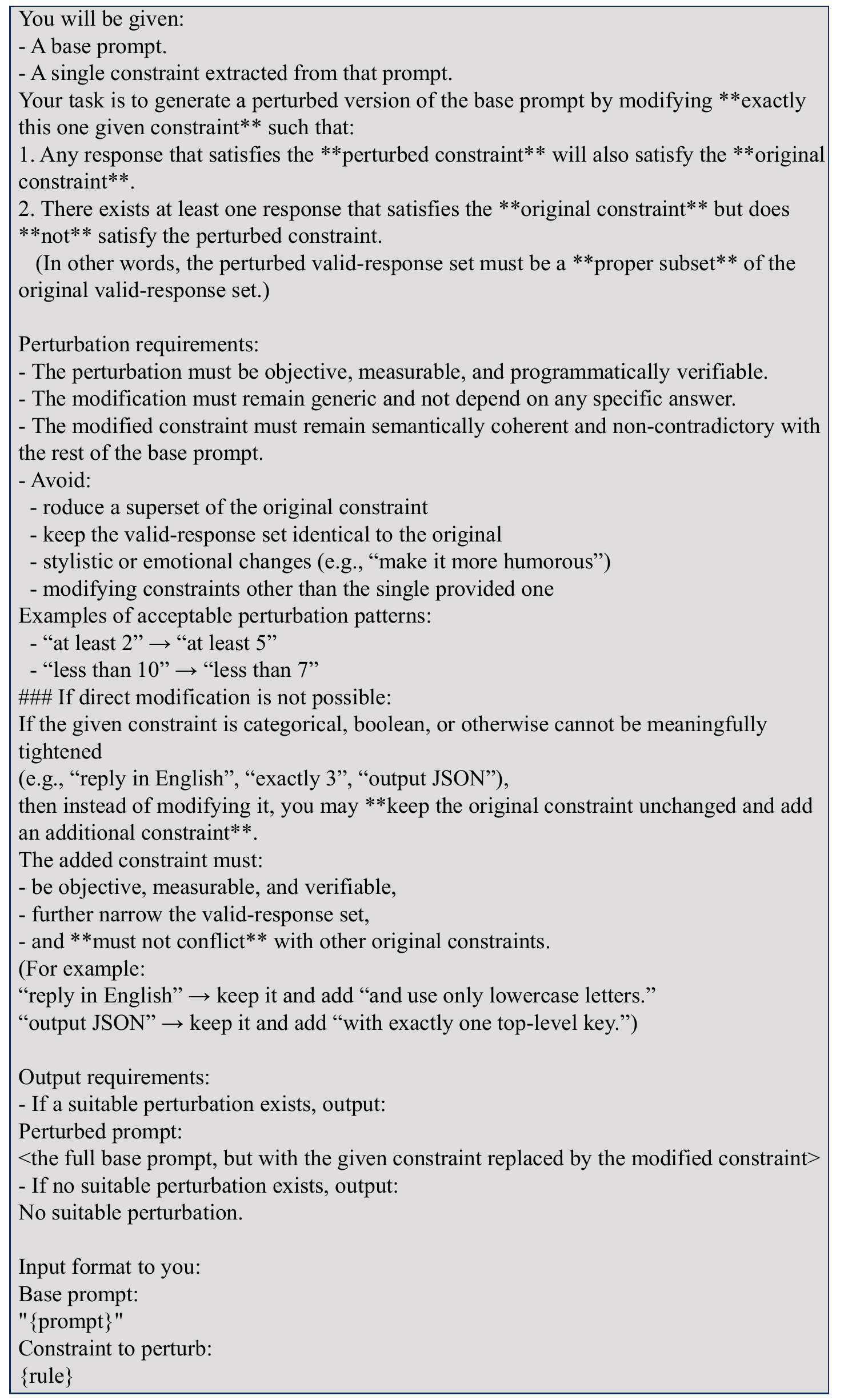}
  \caption{Prompt for the disjoint relationship perturbation.}
  \label{pert4}
  
\end{figure}

\end{document}